\documentclass[runningheads]{llncs}
\usepackage{graphicx}
\usepackage{amsmath}
\usepackage{amsfonts} 

\begin{document}

\title{Implicit neural representations for joint decomposition and registration of gene expression images in the marmoset brain}

\titlerunning{Image registration and decomposition with implicit networks}

\author{Michal~Byra\inst{1,2,}\thanks{Corresponding author.}, Charissa~Poon\inst{1}, Tomomi~Shimogori\inst{3}, Henrik~Skibbe\inst{1}}

\institute{Brain Image Analysis Unit, RIKEN Center for Brain Science, Wako, Japan \and
Institute of Fundamental Technological Research, Polish Academy of 
Sciences, Warsaw, Poland \and Laboratory for Molecular Mechanisms of Brain Development, RIKEN Center for Brain Science, Wako, Japan \\ \email{michal.byra@riken.jp}}

\maketitle             

\begin{abstract}
We propose a novel image registration method based on implicit neural representations that addresses the challenging problem of registering a pair of brain images with similar anatomical structures, but where one image contains additional features or artifacts that are not present in the other image. To demonstrate its effectiveness, we use 2D microscopy \textit{in situ} hybridization gene expression images of the marmoset brain. Accurately quantifying gene expression requires image registration to a brain template, which is difficult due to the diversity of patterns causing variations in visible anatomical brain structures. Our approach uses implicit networks in combination with an image exclusion loss to jointly perform the registration and decompose the image into a support and residual image. The support image aligns well with the template, while the residual image captures individual image characteristics that diverge from the template. In experiments, our method provided excellent results and outperformed other registration techniques.


\end{abstract}

\section{Introduction}

Image registration is a crucial prerequisite for image comparison, data integration, and group studies in contemporary medical and neuroscience research. In research and clinical settings, pairs of images often show similar anatomical structures but may contain additional features or artifacts, such as specific staining, electrodes, or lesions, that are not present in the other image. This difficulty of finding corresponding structures for automatically aligning images complicates image registration. In this work, we address the challenging problem of the gene expression image registration in the marmoset brain. Brain atlases of gene expression, created using images of brain tissue processed through \textit{in situ} hybridization (ISH), offer single-cell resolution of spatial gene expression patterns across the entire brain \cite{corrales2022single,lein2007genome}. However, accurately quantifying gene expression requires brain image registration to spatially align ISH images to a common atlas space. The diversity of gene expression patterns in ISH images causes variations in visible anatomical brain structures with respect to the template image. ISH microscopy images are also susceptible to tissue processing artifacts, resulting in non-specific staining and tissue deformations.

Traditional pair-wise image registration methods use optimization algorithms to find the deformation field that maximizes the similarity between a pair of images. While several deep learning methods based on convolutional neural networks (CNNs) have been proposed for calculating the deformation field between two images \cite{fu2020deep}, such models typically require large training sets and may suffer from generalization issues when applied to images presenting texture patterns that diverge from the training data. Therefore, classic algorithms, such as Advanced Normalization Tools (ANTs) \cite{avants2011reproducible}, are still preferred as off-the-shelf tools for image registration in neuroscience due to scarce experimental data and the diversity of data acquisition protocols and registration tasks.  Recently, implicit neural representations (INRs) have been utilized for image registration in MRI and CT \cite{sun2022mirnf,wolterink2022implicit}, offering a hybrid approach that connects modern deep learning techniques with per-case optimization as used in classical approaches. INRs are defined on continuous coordinate spaces, making them suitable for registration of images that differ in geometry.

In this work, we propose a novel INR-based framework well-suited to address the challenging problem of gene expression brain image registration. We associate the registration problem with an image decomposition task. We utilize implicit neural networks to decompose the ISH image into two separate images: a support image and a residual image. The support image corresponds to the part of the ISH image that is well-aligned with the registration template image in respect to the texture. On the contrary, the residual image presents features of the ISH image, such as artifacts or texture patterns (e.g. gene expression), which presumably undermine the registration procedure. The support image is used to improve the deformation field calculations. We also introduce an exclusion loss to encourage clearer separation of the support and residual images. The usefulness of the proposed method is demonstrated using 2D ISH gene expression images of the marmoset brain. 

\section{Methods}

\subsection{Registration with implicit networks}

The goal of the pairwise image registration is to determine a spatial transformation that maximizes the similarity between the moving image $M$ and the target fixed template image $F$. INRs serve as a continuous, coordinate based approximation of the deformation field obtained through a fully connected neural network. In this study, as the backbone for our method, we utilized the standard approach to registration with INRs, as described in \cite{sun2022mirnf,wolterink2022implicit}. We used a single implicit deformation network $D$ to map 2D spatial coordinates $\bar{x}\in[-1,1]^2$ of the moving image $M$ to a displacement vector $\Delta \bar{x} \in \mathbb{R}^{2}$. Next, the transformation field was determined as $\Phi(\bar{x})= \bar{x} + \Delta \bar{x}$ and the bilinear interpolation algorithm was applied to obtain the corresponding moved image $T_{\Phi}(M)$. 

To train the deformation network, the following loss function based on correlation coefficients was applied to assess the similarity between the moved image $T_{\Phi}(M)$ and the fixed template image $F$:  

\begin{equation}
    \mathcal{L}_{cc}(F, T_{\Phi}(M)) = \frac{1}{2N} \sum_{\bar{x}}  \biggl( \textrm{NCC}(F, T_{\Phi}(M)) + \textrm{LNCC}(F, T_{\Phi}(M)) \biggl),
    \label{eq:cc}
\end{equation}

\noindent where NCC and LNCC stand for the normalized cross-correlation and local normalized cross-correlation based loss functions averaged over the entire image domain consisting of $N$ elements. NCC was used to stabilize the training of the network, while LNCC ensured good local registration results. Additionally, following the standard approach to INR based registration, we regularized the deformation field based on the Jacobian matrix determinant $|J_{\Phi(\bar{x})}|$ using following equation~\cite{wolterink2022implicit}: 

\begin{equation}
    \mathcal{L}_{reg}(\Phi(\bar{x})) = \frac{1}{N} \sum_{\bar{x}} |1 - |J_{\Phi(\bar{x})}| |. 
\end{equation}

\subsection{Registration guided image decomposition}

\begin{figure}[t]
	\begin{center}
		\includegraphics[width=0.9\linewidth]{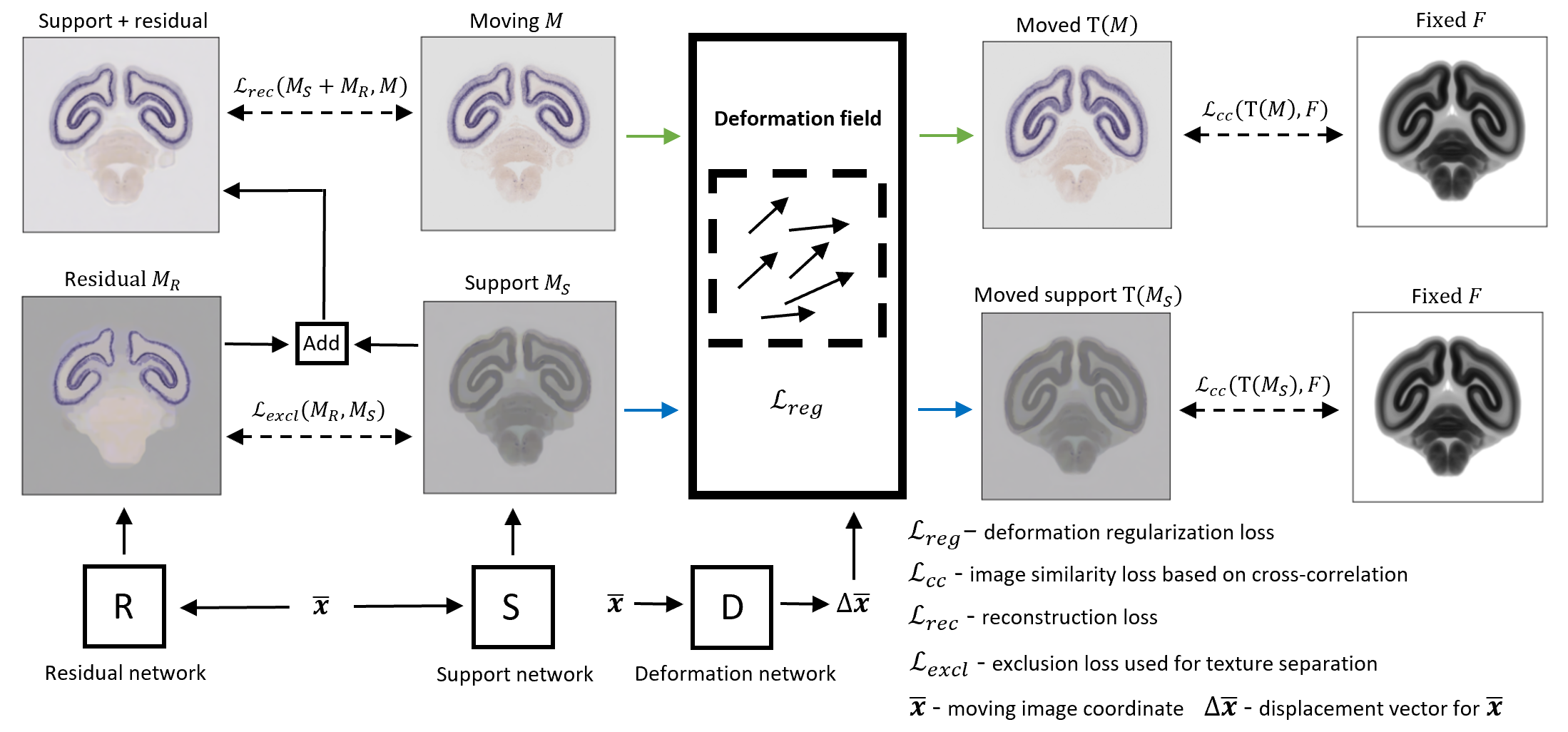}
	\end{center}
	\caption{We use implicit networks $S$ and $R$ to decompose the moving image into the support and residual images. The moving and support images are jointly registered to the fixed template image, which guides the image decomposition procedure to generate a support image that is well-aligned to the fixed image with respect to the texture. The residual image includes the remaining moving image contents that do not contribute to the registration, such as local gene expression patterns or image artifacts.}
	\label{fig:framework}
\end{figure}

Our aim is to improve the registration performance associated with the implicit deformation network $D$. The proposed framework is presented in Fig. \ref{fig:framework}. We assume that the moving image $M$ can be decomposed with separate implicit networks, $S$ and $R$, into two images: the support image $M_S$ and the residual image $M_R$. Ideally, the support image should correspond to the part of the moving image that contributes to the registration performance. On the contrary, we expect the residual image to include image artifacts and texture patterns (e.g. ISH gene expression patterns) that diverge from the fixed template image and undermine the registration procedure. We impose the following condition based on the mean squared error loss function for the decomposition of the moving image:

\begin{equation}
    \mathcal{L}_{rec}(M, M_S+M_R) = \frac{1}{N} \sum_{\bar{x}} (M - M_S - M_R)^2,
\end{equation}

\noindent stating that the support $M_S$ and residual $M_R$ images should sum up to the moving image $M$. To ensure that the support image $M_S$ contributes to the registration with respect to the fixed image $F$, we utilize the cross-correlation based loss function $\mathcal{L}_{cc}(F, T_{\Phi}(M_S))$ (eq. \ref{eq:cc}), where $T_{\Phi}(M_S)$ stands for the transformed support image $M_S$. Therefore, the deformation network is trained to provide the transformation field $\Phi(x)$ both for the moving image and the support image using two cross-correlation based loss functions. This way the training of the deformation network is guided to provide a more detailed transformation field for the contents of the moving image that actually correspond to the fixed template image. Moving image texture patterns that do not correspond to the fixed image have lower impact on the training of the deformation network. 

In practice, it might be beneficial, following INR based methods for obstruction and rain removal, to additionally constrain the image decomposition procedure to obtain more clearly separated support $M_S$ and residual $M_R$ images~\cite{nam2022neural}. For this, we utilize the following exclusion loss to encourage the gradient structure of the implicit networks $S$ and $R$ to be decorrelated \cite{gandelsman2019double}: 

\begin{equation}
 \mathcal{L}_{excl}(M_S, M_R) = \frac{1}{N} \sum_{\bar{x}} \sum_{i,j} |\Gamma (J_{S}(\bar{x}), J_{R}(\bar{x}))|
\end{equation}

\noindent where $\Gamma (J_{M_S}(\bar{x}), J_{M_R}(\bar{x})) = \text{tanh}(J_{S}(\bar{x})) \otimes  \text{tanh}(J_{R}(\bar{x}))$, $\otimes$ indicates element-wise multiplication and indices $i$, $j$ go over all elements of the matrix $\Gamma$. 

In our framework, we jointly optimize all three implicit networks ($D$, $S$  and $R$) using the following composite loss function: 

\begin{equation}
\label{eq:loss}
\begin{split}
    Loss & = \alpha_{1} \mathcal{L}_{cc}(F, T_{\Phi}(M)) + \alpha_{2}\mathcal{L}_{cc}(F, T_{\Phi}(M_S)) + 
 \alpha_{3} \mathcal{L}_{reg}(\Phi(\bar{x})) \\ 
    & +  \alpha_{4} \mathcal{L}_{rec}(M, M_S+M_R) + \alpha_{5} \mathcal{L}_{excl}(M_S, M_R). 
\end{split}
\end{equation}

\noindent The first row of eq. \ref{eq:loss} can be perceived as a standard registration loss, while the second row stands for a regularized image reconstruction loss. 

\subsection{Evaluation}

We designed the proposed method with the aim to address the problem of ISH gene expression image registration. For the evaluation, we used neonate marmoset brain ISH images collected at the Laboratory for Molecular Mechanisms of Brain Development, RIKEN Center for Brain Science, Wako, Japan (gene-atlas.brainminds.jp) \cite{kita2021cellular,shimogori2018digital}. We prepared manual annotations for 2D images from 50 gene expression datasets. Atlas template images were created using ANTs \cite{avants2011reproducible}, based on semi-automatically aligned sets of 2D ISH images from 1942 gene expression datasets. ISH images used to generate the template were converted to gray-scale to meet ANTs requirements and better highlight brain tissue interfaces. 

Performance of the proposed approach was compared to the SynthMorph network and the ANTs SyN registration algorithm based on mutual information metric, as these two methods do not require pre-training and can serve as off-the-shelf registration tools for neuroscience \cite{avants2011reproducible,hoffmann2021synthmorph}. We conducted an ablation study to assess the effectiveness of the proposed representation decomposition approach with and without the exclusion loss. Registration methods were evaluated quantitatively based on Dice scores using manual 2D segmentations prepared for the following five brain structures ranging in size and shape complexity: aqueduct (AQ, 95 masks), hippocampus area (HA, 570 masks), dorsal lateral geniculate (DLG, 370 masks), inferior colliculus (IC, 70 masks) and visual cortex area (VCA, 68 masks).  Segmentations were outlined both for the template and ISH 2D images, resulting in 1114 image pairs corresponding to the same brain regions. We also calculated the percentage of the non-positive Jacobian determinant values to assess the deformation field folding. Moreover, we determined the structural similarity index (SSIM) between the moved images and the template fixed images.  

\subsection{Implementation}

We utilized sinusoidal representation networks to determine the implicit representations \cite{sitzmann2020implicit}. Each network contained five fully connected hidden layers with 256 neurons. We used the Fourier mapping with six frequencies to encode the input coordinates  \cite{tancik2020fourier}. The coordinates and the encoded coordinates were additionally concatenated within the middle layer of the network. Weights of the networks were initialized following the original paper except for the last linear layer of the deformation network $D$, for which we uniformly sampled the weights from [-0.0001, 0.0001] interval to ensure small deformations at initial epochs. Additional details about the network architecture can be found in the supplementary materials. Networks were trained for 1000 epochs using AdamW optimizer with learning rate of 0.0001 on a server equipped with several NVIDIA A100 GPUs \cite{loshchilov2017decoupled}. ISH images of size 360x420 were downsampled to 256x256. Each epoch corresponded to a batch of all image pixel coordinates \cite{sitzmann2020implicit}. After some initial experiments, we set the composite loss function weights (eq.~\ref{eq:loss}) to $\alpha_{1}$=$\alpha_{2}$=$\alpha_{3}$=$\alpha_{5}$=1 and $\alpha_{4}$=100, partially following the previous studies on INRs \cite{nam2022neural,sun2022mirnf,wolterink2022implicit}. The window size for the LNCC loss was set to [32, 32]. Our PyTorch implementation of the proposed INR based registration method is available at https://github.com/BrainImageAnalysis/ImpRegDec. 

\begin{figure}[]
	\begin{center}
		\includegraphics[width=0.8\linewidth]{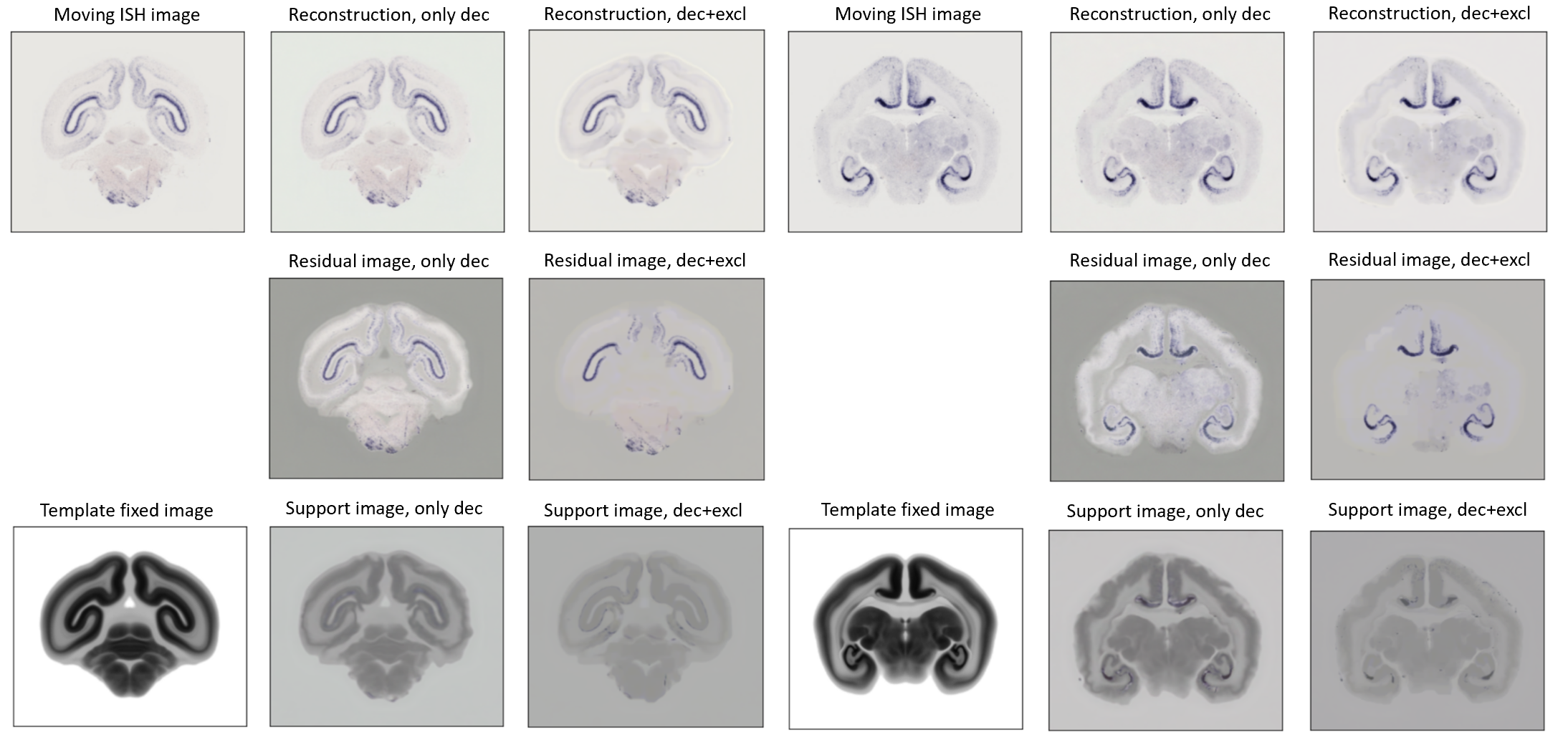}
	\end{center}
	\caption{Illustration of the moving image decomposition obtained with the proposed method (dec). Incorporation of the exclusion loss (excl) resulted in clearer separation of the gene expression texture patterns in the residual images.}
	\label{fig:double_decor}
\end{figure}

\section{Results}

\subsection{Qualitative results}

\begin{figure}[b!]
	\begin{center}
		\includegraphics[width=0.75\linewidth]{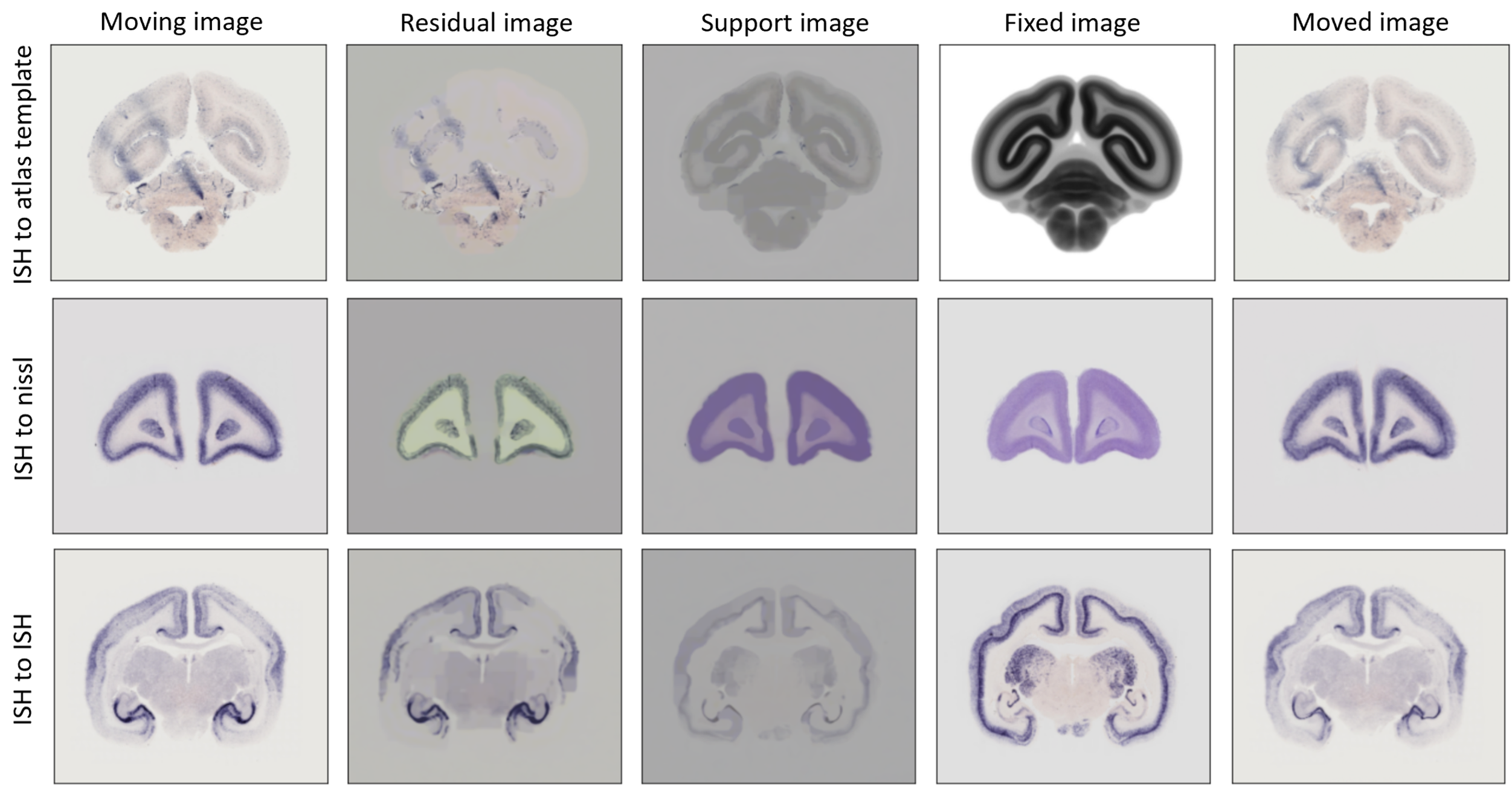}
	\end{center}
	\caption{Proposed technique can be useful for the extraction of microscope image artifacts (e.g. diagonal lines in the first row of images). It can also be applied to register ISH brain images to Nissl images or other ISH images. For such cases the support image presents image style of the fixed image, while the residual image includes local image patterns of the moving image.}
	\label{fig:ish_to_others}
\end{figure}

Support and residual images generated with the proposed method are shown in Fig.\ref{fig:double_decor}. The support images retain the main style and content of the fixed template image, while the residual images include the remaining image contents, along with gene expression patterns not present in the template image. Utilization of the exclusion loss resulted in a clearer and more visually plausible separation between the support and residual images, particularly for gene expression patterns. Fig. \ref{fig:ish_to_others} further highlights the usefulness of the proposed registration guided image decomposition technique. First, our method can be applied to extract microscopy image artifacts, and therefore mitigate their impact on the registration. Second, the proposed method is general and can also be applied to register an ISH gene expression image to a Nissl image. In this case, the color distribution of the support image corresponds to that of a Nissl image, while the residual image presents the local contents of the gene expression image. We also used the proposed method to register an ISH brain image to another ISH image with a different gene expression. For this example, the residual image highlighted the gene expression patterns of the moving image, while the support image showed the gene expression patterns of the fixed image.

\begin{figure}[b!]
	\begin{center}
		\includegraphics[width=0.8\linewidth]{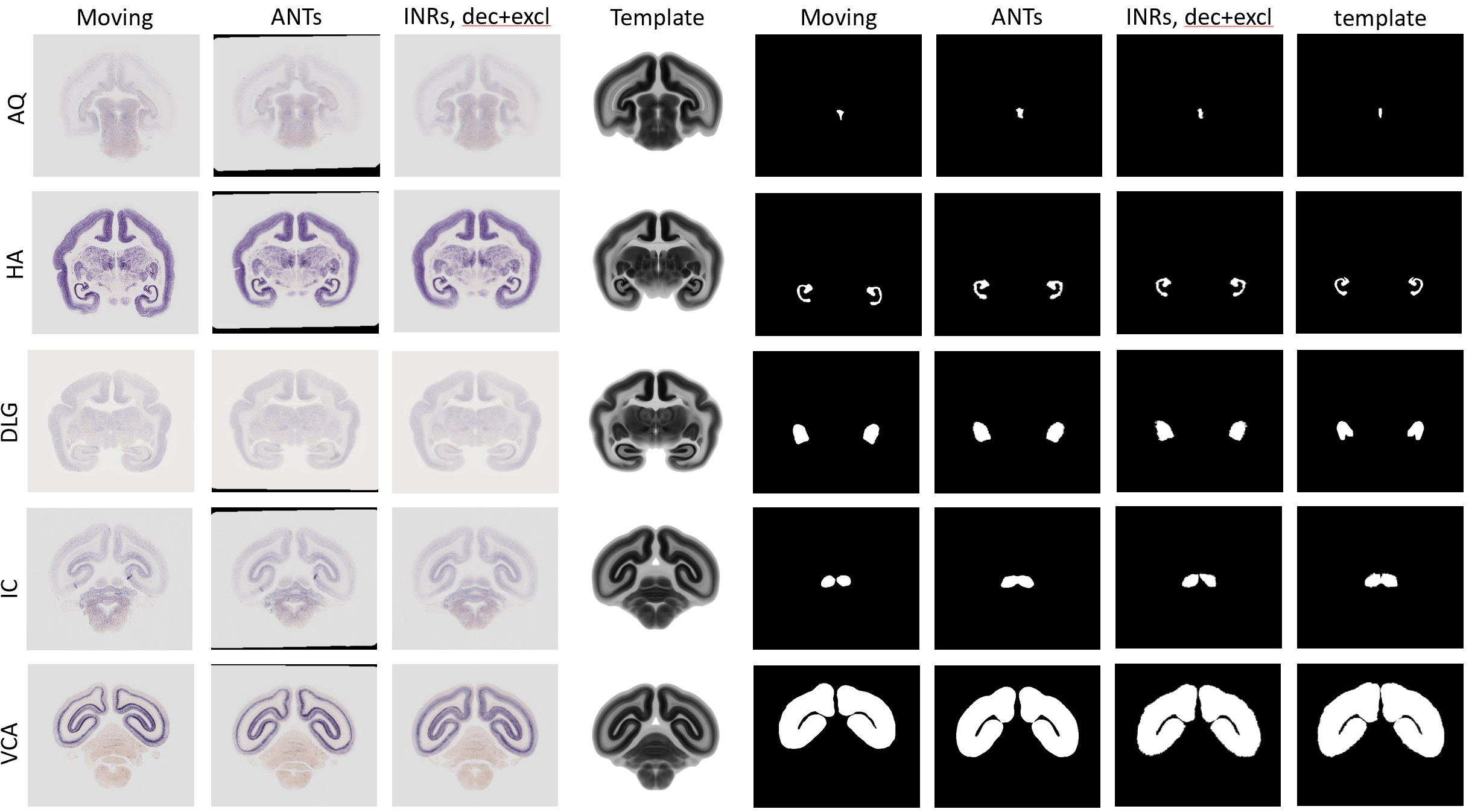}
	\end{center}
	\caption{Comparison of the proposed registration technique based on implicit networks and ANTs. AQ, HA, DLG, IC and VCA indicate the aqueduct, hippocampus area, dorsal lateral geniculate, inferior colliculus and visual cortex area, respectively.}
	\label{fig:segm}
\end{figure}

Fig. \ref{fig:segm} visually compares the registration performance of the proposed technique, equipped with the exclusion loss, to ANTs. We found that the proposed method provided good results both in respect to the image registration and the transformation of the manual segmentations. 

\subsection{Quantitative results}

\begin{table}[t!]
\small
\centering
\caption{Dice scores (mean$\pm$std) determined for the aqueduct (AQ), hippocampus area (HA), dorsal lateral geniculate (DLG), inferior colliculus (IC) and visual cortex are (VCA). Best results are shown in bold. dec and excl stand for the proposed image decomposition technique and the exclusion loss. }
\label{tab:t1}
\begin{tabular}{|c|c|c|c|c|c|}
\hline

Method  & AQ $\uparrow$ & HA $\uparrow$ & DLG $\uparrow$ & IC $\uparrow$ & VCA $\uparrow$  \\ \hline

None &  0.497$\pm$0.194  & 0.311$\pm$0.132   &  0.612$\pm$0.141   &  0.742$\pm$0.169    &  0.848$\pm$0.051    \\ \hline

ANTs SyN &  0.673$\pm$0.102  &  0.644$\pm$0.141   & 0.757$\pm$0.130  &  0.831$\pm$0.133 & \textbf{0.941}$\pm$0.015  \\ \hline

SynthMorph &  0.625$\pm$0.129  &  0.503$\pm$0.190   &  0.719$\pm$0.146  &  0.798$\pm$0.157    &  0.922$\pm$0.034   \\ \hline

INRs &  0.734$\pm$0.071  &  0.657$\pm$0.127   &  0.756$\pm$0.130   &  0.804$\pm$0.191   &  0.922$\pm$0.022   \\ \hline

INRs, dec &  0.748$\pm$0.067  &  0.662$\pm$0.138   & \textbf{0.767}$\pm$0.125    &   0.839$\pm$0.142   &  0.916$\pm$0.033     \\ \hline

INRs, dec+excl   & \textbf{0.749}$\pm$0.063   &  \textbf{0.665}$\pm$0.134   &  0.766$\pm$0.128   &   \textbf{0.845}$\pm$0.143   &   0.920$\pm$0.017    \\ \hline

\end{tabular}

\end{table}

Table \ref{tab:t1} shows Dice scores obtained for the selected marmoset brain regions. Registration techniques based on INRs outperformed the other methods on four out of five brain regions. ANTs achieved better registration results for only one structure, the VCA, which was the largest among the annotated brain regions and already similar in unregistered images with an initial Dice score of 0.848. Additionally, the Dice score for the VCA was high and comparable across all investigated registration methods. Our approach achieved significantly better Dice scores compared to the standard INRs for AQ, HA, DLG and IC  (t-test's $p$-values$<$0.05). Furthermore, incorporating the exclusion loss slightly improved the Dice scores for three structures. 

\begin{table}[t!]
\small
\centering
\caption{Structural similarity index (SSIM) and the percentage of the non-positive Jacobian determinant values (mean$\pm$std) calculated for the investigated registration methods. Best results are shown in bold. dec and excl indicate the proposed image decomposition technique and the exclusion loss, respectively. 
}
\label{tab:t2}
\begin{tabular}{|c|c|c|}
\hline

Method  & SSIM $\uparrow$ & $|J_{\Phi}|\le0$ {[}\%{]} $\downarrow$   \\ \hline

None  & 0.619$\pm$0.046 & ---   \\ \hline

ANTs  & 0.656$\pm$0.059 & $<$\textbf{0.001}   \\ \hline

SynthMorph  & 0.683$\pm$0.039 & $<$\textbf{0.001}    \\ \hline

INRs  & 0.713$\pm$0.052 & 0.353$\pm$0.459    \\ \hline

INRs, dec  & 0.725$\pm$0.054 & 0.359$\pm$0.415    \\ \hline

INRs, dec+excl  & \textbf{0.727}$\pm$0.054 & 0.429$\pm$0.460    \\ \hline

\end{tabular}

\end{table}

SSIM values in Table \ref{tab:t2} show that the registration based on implicit networks provided the most structurally similar results to the template images. With respect to the SSIM metric, our method significantly outperformed other approaches (t-test's $p$-values$<$0.05). ANTs and SynthMorph provided  smoother deformation fields compared to the implicit networks, with significantly lower percentage of folding (t-test's $p$-values$<$0.05). However, the percentage of the folding obtained for the implicit networks was small and acceptable, as defined by folds in 0.5\% of all pixels \cite{qiu2021learning}. The main disadvantage of the proposed approach was the relatively long optimization time of about 90 seconds for a single pairwise registration, resulting from the requirement to jointly train three implicit networks. 

\section{Conclusion}
Our approach based on implicit networks and registration-guided image decomposition has demonstrated excellent performance for the challenging task of registering ISH gene expression images of the marmoset brain. The results show that our approach outperformed pairwise registration methods based on ANTs and SynthMorph CNN, highlighting the potential of INRs as versatile off-the-shelf tools for image registration. Moreover, the proposed registration-guided image decomposition mechanism not only improved the registration performance, but also could be used to effectively separate the patterns that diverge from the target fixed image. In the future, we plan to investigate the possibility of using image decomposition for simultaneous registration and pattern segmentation, and methods to speed up the training  \cite{mehta2021modulated}. We also plan to extend our technique to 3D and test it on medical images that include pathologies.  
\\
\\
\noindent \textbf{Acknowledgement.} The authors do not have any conflicts of interest. This work was supported by the program for Brain Mapping by Integrated Neurotechnologies for Disease Studies (Brain/MINDS) from the Japan Agency for Medical Research and Development AMED (JP15dm0207001) and the Japan Society for the Promotion of Science (JSPS, Fellowship PE21032).

\bibliographystyle{splncs04}
\bibliography{references}

\newpage
\appendix

\section{Network architecture}

\begin{figure}[]
	\begin{center}
		\includegraphics[width=0.99\linewidth]{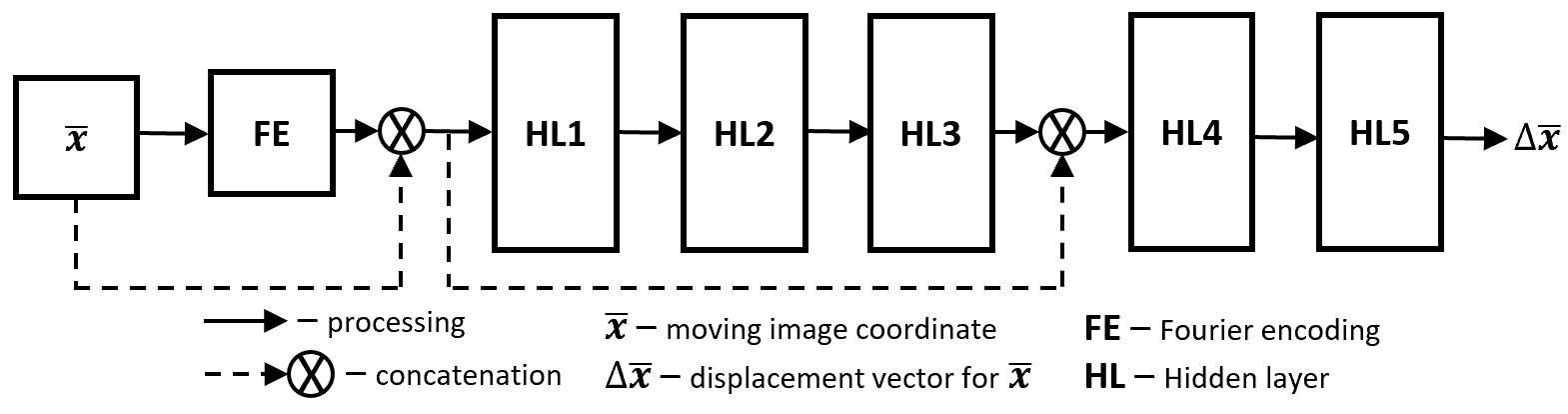}
	\end{center}
	\caption{Architecture of the implicit network used in this study for joint brain image registration and decomposition.}
	\label{fig:network}
\end{figure}

Implicit sinusoidal representation network (SIREN) utilized in our work is depicted in  Fig. \ref{fig:network} and has the following form:

\begin{equation}
\begin{split}
&z^{(0)}= [\bar{x}, \text{FE}(\bar{x})], \\
&z^{(l)} = 
    \begin{cases}
      \rho \left( W^{(l)} z^{(l-1)} - b^{(l)}  \right), & l \in \{1,...,L-1\} \setminus l_{mid} \\
      \left[\rho \left( W^{(l)} z^{(l-1)} - b^{(l)} \right), z^{(0)} \right], & l = l_{mid}
    \end{cases} \\   
&\Delta \bar{x} = W^{(L)} z^{(L-1)} - b^{(L)}, \\
\end{split}
\label{eq: logical 1}
\end{equation}

\noindent where $\bar{x}$ and $\Delta \bar{x}$ stand the moving image coordinate and the corresponding displacement vector, respectively. $W^{(l)}$, $b^{(l)}$ and $z^{(l)}$ correspond to network's weights, bias and post-activation for the $l$-th layer, $l=1,...,l_{mid},...,L$, with $l_{mid}$ indicating the middle layer. Number of the hidden layers was equal to 5 in our work, each consisting of 256 units. Network utilized sine activation function $\rho(y) = \text{sin}(\omega y)$ with $\omega$ standing for the frequency related parameter and set to 30 \cite{sitzmann2020implicit}. FE$(\bar{x})$ indicates the positional Fourier encoding that was concatenated with the input coordinate $\bar{x}$. Additionally, we formed a residual connection by concatenating the image coordinate $\bar{x}$ and the FE$(\bar{x})$ within the middle layer of the network. The positional encoding FE$(\bar{x})$ had the following form \cite{tancik2020fourier}:

\begin{equation}
    \text{FE}(\bar{x}) = [..., \text{cos}(2 \pi \sigma^j \bar{x}), \text{sin}(2 \pi \sigma^j \bar{x}),...] 
\end{equation}

\noindent for $j=0,...,N-1$ and $N$ and $\sigma$ equal to 6 and 2 in our work, respectively. 

Regarding the weight initialization, we followed the SIREN paper \cite{sitzmann2020implicit}. Weights of the network were sampled from the uniform distribution $\mathcal{U}\left( -\sqrt{ \frac{c}{n\omega^2} }, \sqrt{ \frac{c}{n\omega^2} } \right)$, with $c$, $\omega$ and $n$ equal to 6, 30 and 256, respectively. However, for the last layer of the network we sampled the weights from $\mathcal{U}(-0.0001, 0.0001)$ to ensure small displacement vectors $\Delta \bar{x}$ at the initial training epochs.

\end{document}